\documentclass[10pt,twocolumn,letterpaper]{article}

\usepackage[pagenumbers]{iccv} 
\catcode`\_=12
\definecolor{iccvblue}{rgb}{0.21,0.49,0.74}
\usepackage[pagebackref,breaklinks,colorlinks,allcolors=iccvblue]{hyperref}

\def\paperID{*****} 
\def\confName{ICCV}
\def\confYear{2025}

\title{LLMControl: Grounded Control of Text-to-Image Diffusion-based Synthesis with Multimodal LLMs}

\author{Jiaze Wang ~~Rui Chen ~~ Haowang Cui\\
Tianjin University\\
{\tt\small jiaze_w, ruichen, haowangcui@tju.edu}
}

\begin{document}
\maketitle
\begin{abstract}
Recent spatial control methods for text-to-image (T2I) diffusion models have shown compelling results. However, these methods still fail to precisely follow the control conditions and generate the corresponding images, especially when encountering the textual prompts that contain multiple objects or have complex spatial compositions. In this work, we present a LLM-guided framework called LLM\_Control to address the challenges of the controllable T2I generation task. By improving grounding capabilities, LLM\_Control is introduced to accurately modulate the pre-trained diffusion models, where visual conditions and textual prompts influence the structures and appearance generation in a complementary way. We utilize the multimodal LLM as a global controller to arrange spatial layouts, augment semantic descriptions and bind object attributes. The obtained control signals are injected into the denoising network to refocus and enhance attention maps according to novel sampling constraints. Extensive qualitative and quantitative experiments have demonstrated that LLM\_Control achieves competitive synthesis quality compared to other state-of-the-art methods across various pre-trained T2I models. It is noteworthy that LLM\_Control allows the challenging input conditions on which most of the existing methods fail.
\end{abstract}    
\section{Introduction}
\begin{figure}
    \centering
    \includegraphics[width=1\linewidth]{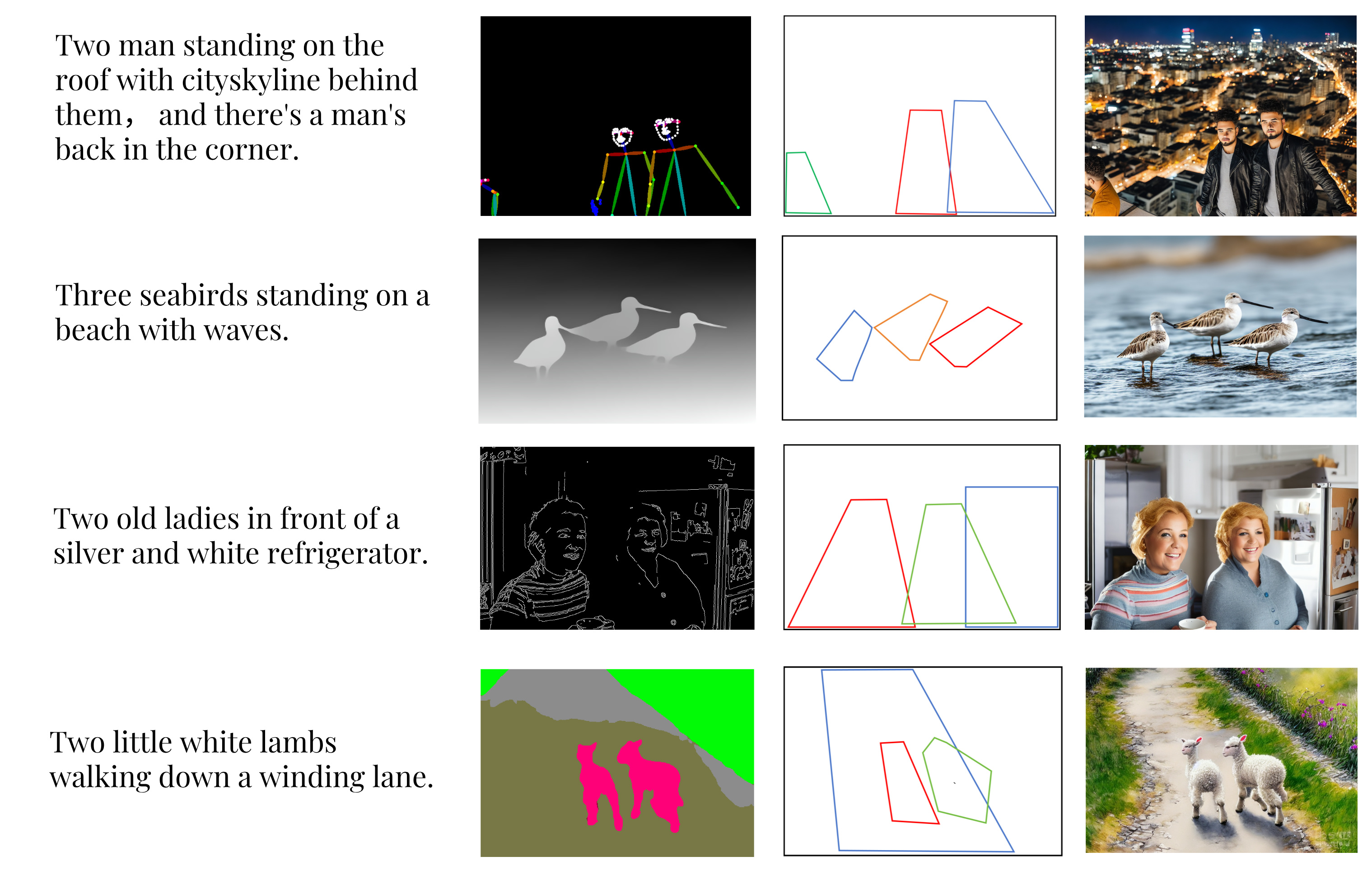}
    \caption{The figure shows the results generated by our method. The text are the complex prompts used in the generation. the left column are the spatial image conditions, the middle column are the polygon layouts generated by our method, and the right column are the generated images.
}
    \label{fig:1}
\end{figure}

\section{Introduction}
\label{sec:intro}

With the emergence of a range of diffusion models \cite{Alpher39,Alpher40,Alpher41} represented by Stable Diffusion \cite{Alpher38,Alpher15,Alpher42}, T2I task has taken on a whole new look. Using T2I diffusion models, we can now generate high-quality images that meet mental expectations with simple inputs. Diffusion models generate results by iteratively transforming an initial simple distribution into a complex high quality distribution. 

However, there are still some problems in the T2I diffusion generation task. For users, it is difficult to accurately describe the desired effect by text alone and text that is too complex and lengthy is sometimes difficult to be directly understood by diffusion models. On top of the underlying model, more and more works \cite{Alpher03,Alpher04,Alpher05,Alpher06,Alpher07,Alpher08,Alpher14,Alpher26} are introducing additional spatial image to achieve more precise control. But we find that the trained model tends to prioritize image-space conditions over textual descriptions, probably because the input-output image pairs are tightly spatially aligned, which often leads to the results that are not aligned with the user's textual instructions. Moreover, while methods such as ControlNet \cite{Alpher03} and T2I Adapter \cite{Alpher08} allow users to provide more precise spatial control of pre-trained T2I diffusion models by providing various types of image conditions as well as textual descriptions, these methods \cite{Alpher9,Alpher37,Alpher20,Alpher08} often require the training of different add-on modules for different types of spatial images, leading to unnecessary repetitive training and expensive computational costs.
\begin{figure*}
    \centering
    \includegraphics[width=1\linewidth]{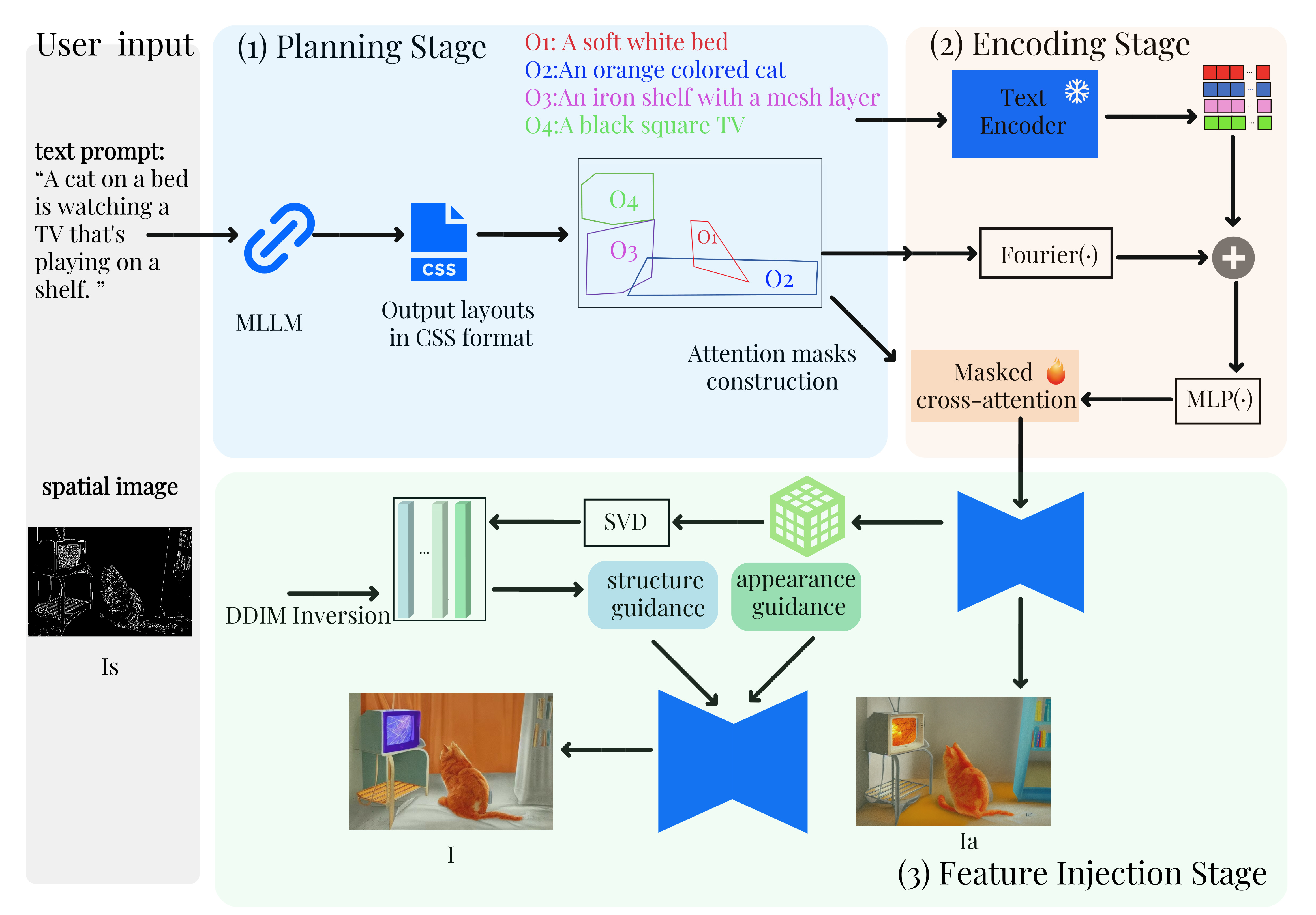}
    \caption{The figure shows the architecture of our approach. Our method consists of three stages. Firstly, we utilize MLLM planning to generate tasks. Then, we encode to get the embedding and generate cross-attention masks. Finally, we utilize structure guidance and appearance guidance to generate the final image.
}
    \label{fig:2}
\end{figure*}

To address the above limitations, we propose LLMControl, a controllable T2I diffusion method. Previous work \cite{Alpher28,Alpher43,Alpher44,Alpher45} has shown that diffusion models usually predict low-frequency content at the beginning of the denoising process and high-frequency details at the final stage of the process. Low-frequency semantics correspond to main objects and locations, while high-frequency semantics are related to detailed attributes. So we will use multimodal LLM (MLLM) \cite{Alpher56,Alpher57,Alpher58,Alpher59,Alpher60} as a global task planner to guide the generation process through three levels of layout, structure and appearance with user input of textual prompts and spatial image conditions. Some of our generated image results are shown in Figure \ref{fig:1}.

Specifically, LLMControl has three stages, As shown in figure \ref{fig:2}.  In the Planning Stage, MLLM can decompose the generation task into a combination of subtasks by splitting the complex text input into multiple objects, each with its own appearance description and layout. The results of subtasks will eventually be combined into the final image. This is followed by an Encoding Stage, where the layout and appearance description of each object will be encoded as embeddings and connected together, and then fed into the diffusion model's cross-attention. In contrast to previous work \cite{Alpher13,Alpher24,Alpher05,Alpher33,Alpher11,Alpher12} that used rectangles as layouts, we propose a new visual representation that describes the position of objects with polygons. Finally, there is a Feature Stage, where we obtain diffusion features of the spatial image by DDIM Inversion \cite{Alpher23}, which will be used for structure guidance. 

Notably, we use pre-trained MLLMs and diffusion models to reduce training costs. Our approach is also compatible with different types of spatial images without the need to train specific components for different additional image inputs. LLMControl performs well under challenging input conditions. It obtains competitive image synthesis quality while providing stronger image-text alignment and supporting a broader set of control signals. In summary, our contributions include:

 1) In this paper, we proposed a new controllable T2I generation method, LLMControl, which guides the generation process by hierarchically utilizing layout, structure and appearance to achieve precise alignment and high level of image quality.

2) We proposed a new visual primitive representation, i.e., each object in an image is represented by an irregular polygonal layout with textual descriptions. This preserves richer and more precise visual information, and also facilitates human and MLLM understanding and construction.

3) We extracted structural features of the spatial image condition and injected them during the generation process, thus guiding the generated image to follow the structure of the spatial image. The spatial image conditions of various modalities have a unified representation under our design without having to train multiple components

4) We have conducted extensive experiments on various benchmarks, showing that LLMControl is comparable to previous SOTA methods.

\section{Related work}
\label{sec:rela}
\subsection{T2I generation models}
Compared to GAN \cite{Alpher46,Alpher47,Alpher48} and autoregressive models \cite{Alpher49,Alpher50,Alpher51}, diffusion models \cite{Alpher30,Alpher39,Alpher40,Alpher41,Alpher23} have become more popular recently due to the high quality and diversity of the images they generate. So far, many image generation methods based on diffusion models have been proposed. Among them, Stable Diffusion \cite{Alpher38,Alpher15,Alpher42} is popular due to its strong performance and robust open-source status. It usually employs the latent diffusion model (LDM) , which allows the denoising process to operate in a low-dimensional latent space, thus reducing the computational cost. Besides, a number of fine-tuning techniques \cite{Alpher04,Alpher07,Alpher9,Alpher17,Alpher27,Alpher10,Alpher45,Alpher20,Alpher35,Alpher34} have been proposed that enhance the ability of T2I diffusion models to reduce training costs and improve task adaptation. These models have also been extended for other image processing tasks such as image editing, image restoration, and style conversion.

\subsection{Diffusion model meets LLM}
LLMs have strong comprehension, reasoning, and generalization capabilities \cite{Alpher52,Alpher53,Alpher54,Alpher55}. So more and more studies are utilizing LLMs to assist T2I tasks. Previous works \cite{Alpher25,Alpher24} take the advantage of of LLMs to segment and refine the generation task, reflecting the planning ability of LLMs. Some works \cite{Alpher11,Alpher13,Alpher25,Alpher01,Alpher33} have also used LLMs to assist diffusion models in understanding spatial location relationships. LLMs have transformed a variety of NLP tasks with special generalization capabilities. Our work, however, further utilizes the cross-modal capabilities of the MLLM to further interactively align textual prompts and spatial image conditions, thus enhancing the compliance of generated results to user's instructions.

\subsection{Spatial image conditions}
For more precise control, spatial image conditions (\eg repaired masks, sketches, keypoints, depth maps, segmentation maps, layout maps, or even a complete image) have been used to guide the generation process, and they are often interacted with the user's textual prompts in the diffusion model's cross-attention. In order to better utilize the additional image guidance, some works have devised clever fine-tuning methods for image generation, with Dreambooth \cite{Alpher04} and HyperDreambooth \cite{Alpher07} requiring only three or four images to fine-tune the T21 diffusion model for theme-driven generation. Other works \cite{Alpher08,Alpher16,Alpher03} have trained new components or neural network layers while retaining the original parameters of the model, enhancing compatibility with various types of spatial images. Composer \cite{Alpher18,Alpher37} and others, on the other hand, require the diffusion model to be trained from scratch, which undoubtedly increases the computational cost.

\begin{figure}
    \centering
    \includegraphics[width=1\linewidth]{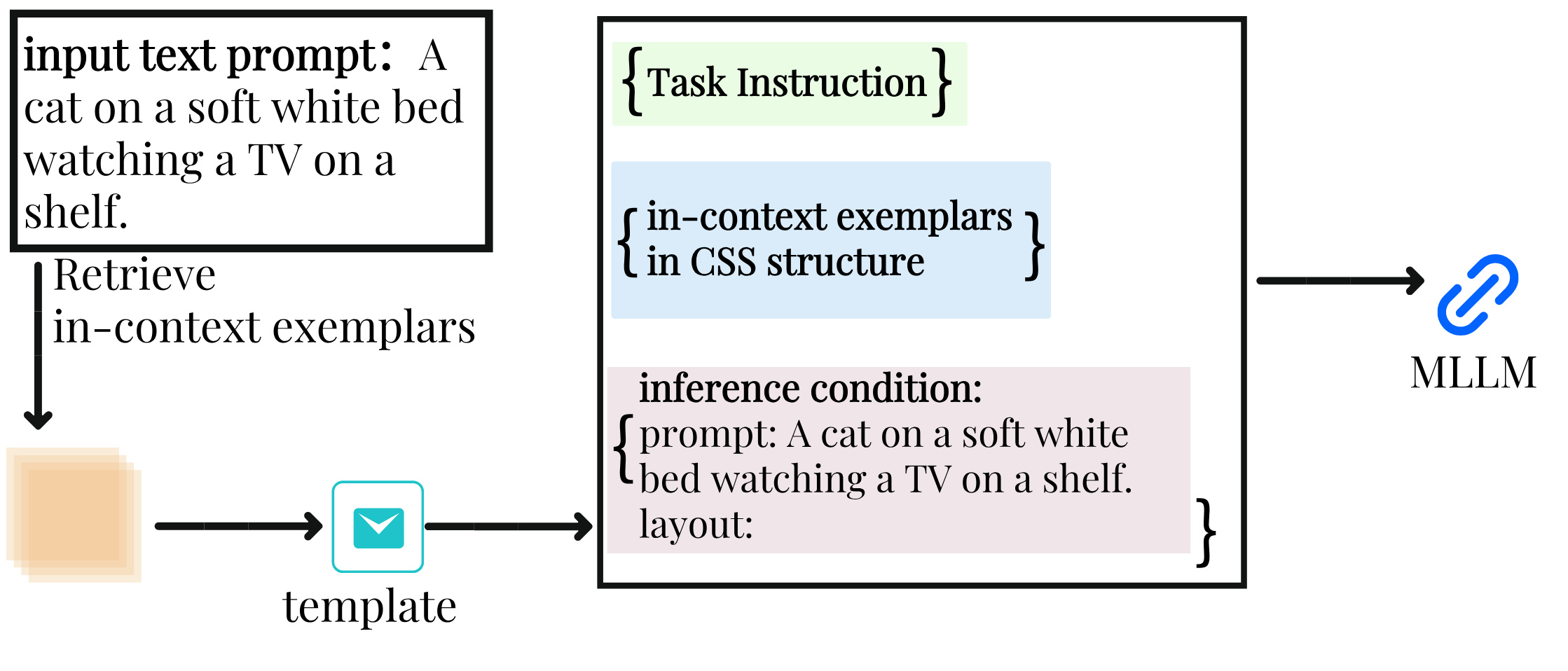}
    \caption{The figure shows  the template construction. Specifically, it consists of three parts: the task instruction describes the task of the LMM in natural language. the In-context examplars provide similar examples for the LLM to learn from by indexing the results.. Inference conditions are user-input text prompts.
}
    \label{fig:enter-label}
\end{figure}

\section{Methods}
\subsection{Visual Representation Based on Path Clipping.}
Given an image containing multiple objects, we would like to extract an object-level representation as a visual primitive. The primitives should meet the following conditions: 1) They should contain as many fine-grained details as possible, so as to ensure that they can be reconstructed efficiently during the diffusion generation process. 2) The Representation should be human-friendly, and they should be easy to interpret and manipulate, so that the user can create and edit them efficiently. 3)They should be represented in a way that is easy for MLLM to understand, thus enhancing MLLM's interpretation of the spatial meaning implicit in the primitives. To this end, we propose a new representation of visual primitives called Path Clip, where each representation describes a single object in the scene. The representation consists of two components: the path parameters and the appearance description of the text.
\medskip

 \noindent\textbf{Path parameters.} The path parameters firstly create a rectangular box of specified size and position using a vector \(\left[ {{c_x},{c_y},w,h} \right]\), where \(\left( {{c_x},{c_y}} \right)\)  is the center point of the rectangle and \(w\) and \(h\) are the width and height. Based on this, we add some clipping point location parameters \(\left( {{x_i},{y_i}} \right)\) so that the rectangular box is clipped to obtain an polygon. We follow previous works \cite{Alpher13} in using CSS format to represent path parameters \(\tau : = \left[ {{c_x},{c_y},w,h,{x_i},{y_i},...,{x_k},{y_k}} \right]\), so that LLM can better understand their spatial meaning. Each layout structure we generate in the context example starts with a category name followed by the declaration part in the CSS style, which is " \texttt{object [cx: ?px, cy: ?px, w: ?px, h: ?px, clip path: polygon(?px ?px, ?px ? px,......)]". }We use polygons instead of rectangles or ellipses as bounding boxes, which allows a more precise description of the shape and size of objects. 
 \medskip

\noindent\textbf{Appearance description. }Appearance description is text statements that describes the visual appearance of the objects within the polygonal box. For ease of understanding, we begin with the category name and use natural language for the appearance description (\eg, color, texture, material).

\begin{figure*}
    \centering
    \includegraphics[width=1\linewidth]{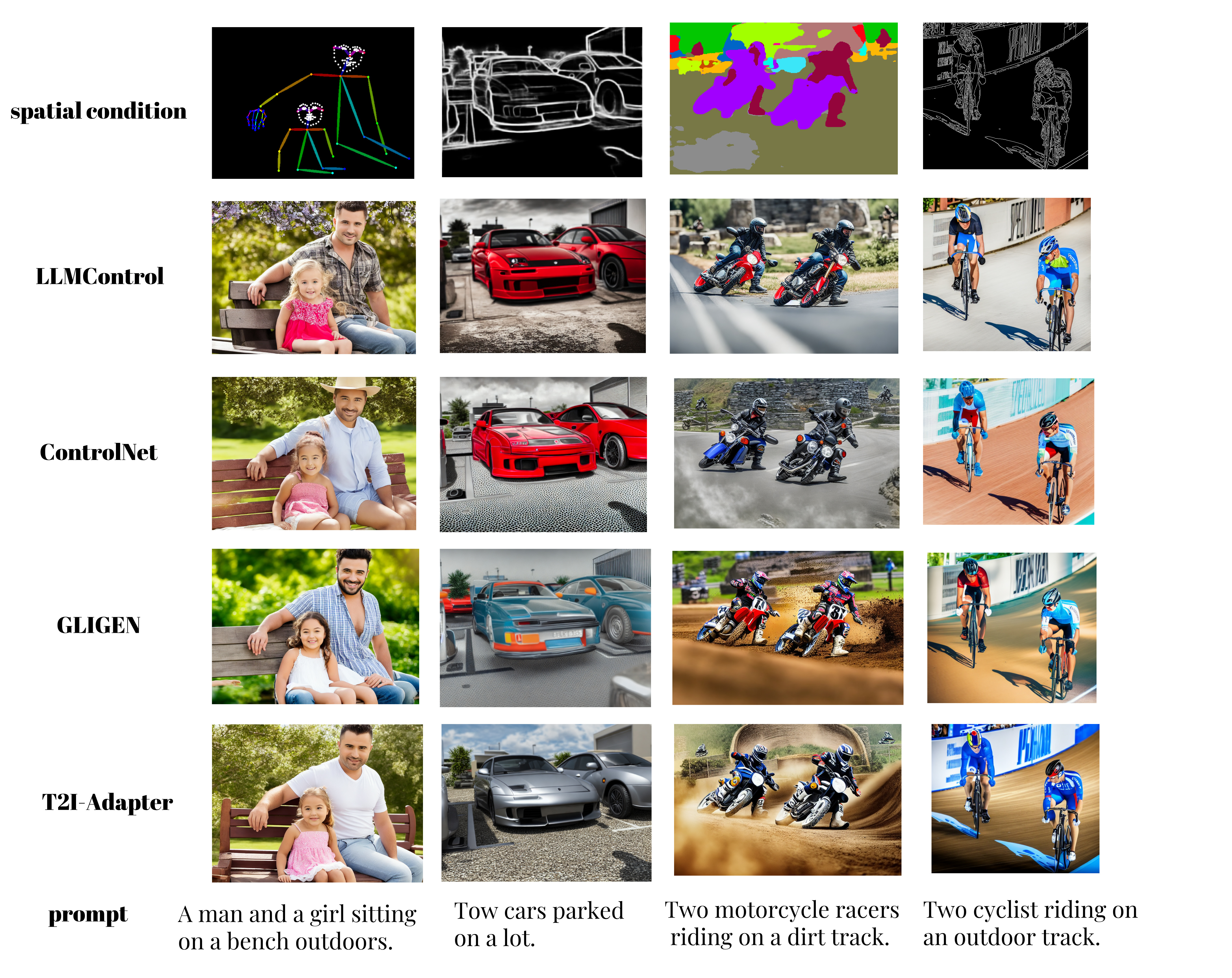}
    \caption{The figure shows how our method compares with other methods in terms of controllability. Inputting the same spatial image condition and textual prompt on different models, our generated images are of better quality and more consistent with the textual prompt.
}
    \label{fig:4}
\end{figure*}

 \subsection{Alignment of layout to description using masks}
The path parameters \(\tau \) contains the location information of the object, we first use the Fourier feature encoding \cite{Alpher61} to get the path embedding \({e_\tau } = Fourier\left( \tau  \right) \in {\mathbb{R}^{{d_\tau }}}\). We denote the appearance description as \(\theta : = \left[ {{s_1},{s_2},...,{s_L}} \right]\). We then use the CLIP text encoder \(E\left(  \cdot  \right)\) to encode textual appearance descriptions to obtain appearance embedding \({e_\theta } = E\left( \theta  \right) \in {\mathbb{R}^{L \times {d_\theta }}}\). Subsequently, we connect them  so that the spatial location of the object is bounded to the appearance description, which is helpful in avoiding object loss or overlap when generating objects containing descriptions with similar appearance descriptions. The final embedding is denoted as :
\begin{equation}
[{e_b} = MLP\left( {\left[ {{e_{\pi 1}},{e_{\pi 2}},...,{e_{\pi L}}} \right]} \right)\
,\label{eq:1}
\end{equation}where \({e_{\pi l}}: = [{e_{\theta l}};{e_\tau }] \in {\mathbb{R}^{{d_s} + {d_\theta }}}\) and \(l \in \left\{ {1,2,...L} \right\}\), meaning that the two embeddings will be connected along the feature dimension. \(MLP\left(  \cdot  \right)\) is an MLP neural network. Text and image features interact at the cross-attention layer of the diffusion model, a process that can be viewed as coloring the feature structures extracted from the spatial image condition, with the coloring scheme following the descriptions extracted from the textual prompts. The standard cross-attention is:
\begin{equation}
CA\left( {Q,K,V} \right) = \sigma \left( {\frac{{Q{K^T}}}{{\sqrt d }}} \right)V\
,\label{eq:2}
\end{equation}where \(Q\), \(K\) and  \(V\) represents the Query, Key and Value in attention layers. \(\sigma \left(  \cdot   \right)\) is the softmax function. We observe that , each visual primitive containing an object focuses on each feature “pixel” of the feature image, which can confuse the model. This is because in reality a visual primitive only determines the coloring scheme for local locations. Therefore, we use masked cross-attention. We generate a mask \({M^i}\) based on the polygonal border, with values set to \(1\) in the polygonal region and \(0\) in the remaining locations. We use \({M^i}\) to mask the feature image so that each primitive embedding focuses only on the corresponding polygonal region. The masked cross attention is defined as:
\begin{equation}
CA_m^i = \sigma \left( {\frac{{Q{K^i}^{^T}}}{{\sqrt d }}} \right){V^i} \cdot {M^i}
,\label{eq:3}
\end{equation}where \({K^i}\) and \({V^i}\) are obtained from \({e_b}\) , and \(Q\) is obtained from the image feature map.

\subsection{SVD-based feature extraction and injection}
With the two stages described above, we have been able to generate the desired image based on the layout and appearance descriptions as many previous works did. However, we still lack control over the structure. On the one hand, we have difficulty in describing specific structures with simple text. On the other hand, complex spatial structures are difficult to be understood and structured by MLLM. For this reason we have designed a module for structural feature injection to allow the diffusion model to receive guidance from additional spatial images \({I_s}\) when generating the target image \(I\). However, simply injecting the features will cause appearance leakage, i.e., the appearance details of the generated image mimic the spatial image conditions. This problem is especially serious when injecting in some deep layers. For this reason we employ the following two strategies to mitigate the problem.
\medskip

\noindent\textbf{\textbf{Self-attention layer injection.}} For the injection location, we chose the self-attention layer. Self-attention is another important attention mechanism in the diffusion model besides cross-attention. Self-attention represents the affinity between spatial features. This means that it preserves fine layout and shape details, so we can characterize the structure of the image with keys and queries of self-attention.
\medskip

 \noindent\textbf{\textbf{SVD on Diffusion Features.}} If we directly inject the weight matrix of self-attention as diffusion features, this still does not mitigate the appearance leakage, which may be due to overfitting. For this reason, we perform singular value decomposition (SVD) on the matrix consisting of diffusion feature vectors. For a matrix \(W\), its SVD is \(W = U\sum V\), and \(\sum  = diag\left( \sigma  \right)\), where \(\sigma  = \left[ {{\sigma _1},{\sigma _2},{\sigma _3}...} \right]\) , are the singular values in descending order. The vectors corresponding to the singular values at the top of the order are strongly correlated with the pose and shape of the object, and thus can be used as base vectors for structural features.

Specifically, guided by the layout and appearance descriptions, we use DDIM sampling \cite{Alpher23} to generate the image \(Ia\), called primitive generation. In this process, we obtain the time-dependent diffusion feature \(\left\{ {F_t^a} \right\}\) from the denoising network. We perform SVD on the matrix of these feature vectors to obtain the time-dependent semantic basis \( \{Bs \}\). We then perform DDIM Inversion on \({I_s}\) to obtain the diffusion features \(\left\{ {F_t^s} \right\}\) and project them onto \(\left\{ {{B_s}} \right\}\) to obtain the structure coordinates \({S_t}\). Combined with the mask \({M^i}\) obtained in the Encoding Stage, we can perform feature injection to accomplish precise structural control for the generation of final image.
\medskip

 \noindent\textbf{Guidance of the generation.} Finally, we will generate the image \( I\). We take a similar design to FreeControl \cite{Alpher01} and add the appearance guidance \({g_a}\) and structure guidance \({g_s}\) to the classifier-free guidance. The original classifier-free guidance \cite{Alpher22} is as follows:
 \begin{equation}
 {{\varepsilon _t }} = \left( {1 + \omega } \right){\varepsilon _\theta }\left( {{X_t};t,c} \right) - \omega {\varepsilon _\theta }\left( {X_t;t} \right)
,\label{eq:4}
\end{equation}
\(\omega \) is the guidance weights and is used to adjust the degree of alignment with condition \(c\). The classifier-free guidance plays a crucial role in enhancing the image-to-text alignment of the generated samples. The energy function \({g_a}\) is then denoted as:
 \begin{equation}
{g_a}\left( {\left\{ {v_t^{\left( k \right)}} \right\};\left\{ {v_k^{a\left( k \right)}} \right\}} \right) = \frac{{\sum\nolimits_{k = 1}^{{N_a}} {\left\| {v_t^{\left( k \right)} - } \right.} \left. {v_t^{a\left( k \right)}} \right\|_2^2}}{{{N_a}}}
,\label{eq:5}
\end{equation}
\({\left\{ {v_t^{\left( k \right)}} \right\}}\) and \({\left\{ {v_k^{a\left( k \right)}} \right\}}\) denote the weighted spatial averages of the diffusion features of \(I \)and \(Ia\). It makes \(I\) mimic the appearance details of \(Ia\) during the generation process.  And \({\left\{ {v_t^{\left( k \right)}} \right\}}\) is defined as: 
 \begin{equation}
v_k^{\left( k \right)} = \frac{{\sum\nolimits_{i,j} {\sigma \left( {{{\left[ {s_t^{\left( k \right)}} \right]}_{ij}}} \right)} {{\left[ {{f_t}} \right]}_{ij}}}}{{\sum\nolimits_{i,j} {\sigma \left( {{{\left[ {s_t^{\left( k \right)}} \right]}_{ij}}} \right)} }}
,\label{eq:6}
\end{equation}

\(i\) and \(j\) are the spatial indices of \({S_t}\) and \({F_t}\).  \(k\) is the channel index of \({\left[ {{S_t}} \right]_{i,j}}\). \(\sigma \left(  \cdot  \right)\)is the sigmoid function. The energy function \({g_{s,f}}\) in forward process is denoted as:

\begin{equation}
{g_{s,f}}\left( {{S_t};S_t^s,M} \right) = \frac{{\sum\nolimits_{i,j} {{m_{ij}}\left\| {\left[ {{s_t}} \right] - \left. {\left[ {s_t^s} \right]} \right\|_2^2} \right.} }}{{\sum\nolimits_{i,j} {{m_{ij}}} }}
,\label{eq:7}
\end{equation}and the energy function \({g_{s,b}}\) in forward process is denoted as: 

\begin{equation}
{g_{s,b}}\left( {{S_t};S_t^s,M} \right) = s\frac{{\sum\nolimits_{i,j} {\left( {1 - {m_{ij}}} \right)\left\| {\max \left( {{{\left[ {{s_t}} \right]}_{ij}} - {\tau _t},0} \right)} \right\|} _2^2}}{{\sum\nolimits_{i,j} {\left( {1 - {m_{ij}}} \right)} }}
,\label{eq:8}
\end{equation}where \(i\) and \(j\) are the spatial indices of \({S_t}\), \({S_t^s}\) and \(M\), and \(s\) is the balance weight. The threshold \({{\tau _t}}\) is the maximum acquisition per channel, which helps to segment the foreground.
Our final score is estimated as:

\begin{equation}
{\widehat \varepsilon _t} = \widetilde \varepsilon  + {\lambda _s}\left( {{g_{s,f}} + {g_{s,b}}} \right) + {\lambda _a}{g_a}
,\label{eq:9}
\end{equation}where \({\lambda _s}\) and \({\lambda _a}\) are the guidance strengths of structure and appearance.

\begin{figure*}
    \centering
    \includegraphics[width=1\linewidth]{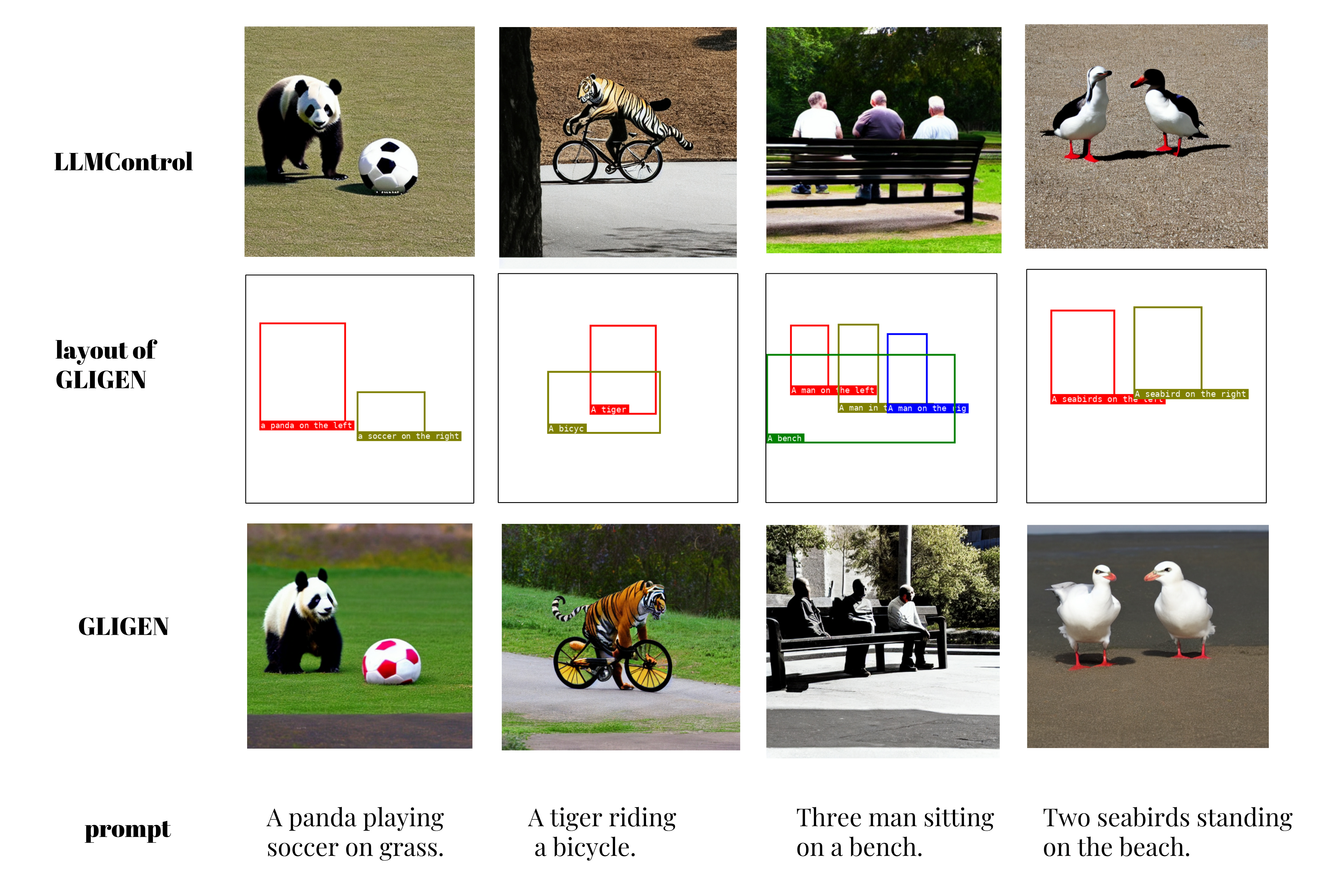}
    \caption{The images show our method with another layout based generation method GLIGEN. The first row shows our generated imageS. The second row shows the layouts generated by GLIGEN. The third row shows the images generated by GLIGEN. It can be seen that our method still produces visually satisfying images in the face of complex textual prompts and spatial structures, while GLIGEN suffers from structural distortion.
}
    \label{fig:5}
\end{figure*}
\section{Experiments}

\subsection{LLM-grounded T2I Generation}

\begin{table}
  \centering
  \begin{tabular}{@{}lcc@{}}
    \toprule
    Methods & $FID \downarrow$ & $CLIP \uparrow$ \\ 
    \midrule
    CogView     & 28.21 & 0.2632 \\
    Make-a-Scene& 15.62 & 0.2911 \\ 
    Imagen      & 7.22  & 0.3114 \\
    GLIGEN      & 12.03 & 0.3013 \\
    ours        & 8.44  & 0.3222 \\
    \bottomrule
  \end{tabular}
  \caption{Evaluation of zero-shot generation quality and layout-guided controllability on MS-COCO validation set. The FID score reflects the visual quality of the generated image, while the CLIP score reflects controllability.}
  \label{tab:1}
\end{table}
\begin{table}
  \centering
  \begin{tabular}{@{}lcccccc@{}}
    \toprule
   Methods & Depth & HED & Normal & Sketch & Canny \\
    \midrule
    ControlNet & 0.314 & 0.320 & 0.321 & 0.317 & 0.319 \\
    P2P & 0.255 & 0.263 & 0.261 & 0.264 & 0.256 \\
  PNP & 0.282 & 0.289 & 0.277 & 0.282 & 0.283 \\
 FreeControl & 0.319 & 0.321 & 0.323 & 0.324 & 0.321 \\
ours & 0.332 & 0.334 & 0.329 & 0.331 & Canny \\\\
    \bottomrule
  \end{tabular}
  \caption{CLIP scores on various types of spatial image conditions. Our method achieved consistently high scores.}
  \label{tab:2}
\end{table}
\textbf{\textbf{Training Strategy.}} We first prepare the image dataset with Path Clip representations needed for training. We use a dataset of random 1 M image-text pairs with high scores in the Common Crawl Web index and adjust the resolution of the images to \(512 \times 512\). Subsequently, we apply ODISE \cite{Alpher36} to obtain the instance segmentation map. To obtain the polygon corresponding to each instance, we use a particle swarm algorithm (PSO) optimization \cite{Alpher62} to maximize the IoU metric for polygon fitting. We set the number of vertices \(k\) between \(4\) and \(6\). We generate the outer join matrix of the polygon by computing the maximum value of the vertex coordinates. At this point, we obtained the complete path parameters. We used LLaVA-1.5 to add subtitles to the instances in each segmentation graph. The captions include the category name, color, texture and material, and also describe the subsections within the object area. 
We train on the latent space, specifically, we map the image onto a \(64 \times 64 \times 4\) latent space. During training, we randomly discard global captions with \(50\%\)
 probability to promote the model's dependence on the Path Clip representation. We use the AdamW optimizer and set the learning rate to \(5 \times {10^{ - 5}}\). In the first \(10 K\)  steps, we use a linear warm-up strategy.\medskip

\noindent\textbf{\textbf{Experiment Setup.}} For the diffusion model we used Stable Diffusion v1.5 and for the MLLM we used GPT3.5-chat and GPT 4. We used the ImageNet-R-TI2I dataset from PnP \cite{Alpher02} as the baseline dataset, which contains \(30\) images from \(10\) object classes. We converted the images to their respective Canny edges, HED edges, sketches, depth maps, and normal maps and used them as spatial image inputs in our experiments. For the feature injection, we use the value of keys from the first self-attention of the U-Net decoder as a diffusion feature. The DDIM sampling takes \(200\) steps and the DDIM Inversion takes \( 1000 \) steps. Structure and appearance guidances are used for the first \(120 \) steps. \({\lambda _s} \in \left( {400,1000} \right]\), \({\lambda _a} = 0.2{\lambda _s}\), \({N_a} = 2\).
\medskip

\noindent\textbf{\textbf{Evaluation metrics. }}For visual quality, we compare the FIDs of the generated and real images. The FID evaluates the quality of the generated model by calculating the distance between the two image distributions. For controllability, we evaluate the alignment using the CLIP score, which is obtained by computing the average cosine similarity between the image embedding and the text embedding of the generated images. Higher scores imply better consistency. To evaluate the generation of layout adherence, we compute the precision, recall, and accuracy of the layout counts and their spatial locations. Specifically, we use Grounding-DINO \cite{Alpher63} to detect each instance, and based on the detected bounding box and ground truth, we compute the precision, recall and accuracy of the generated layout counts and their spatial locations.
\medskip

 \noindent\textbf{\textbf{Quantitative results. }}In Table \ref{tab:1}, our method achieves better FID scores, but still slightly inferior to Imagen. but in terms of consistency, we achieve the highest CLIP scores, which means that our method is able to achieve a high level of image-text alignment. In Table \ref{tab:2}, on the comparison of consistency, our CLIP scores have higher scores compared to other methods. We achieved similar scores on five common spatial image conditions, which also implies that our method is suitable for receiving diverse spatial image structures.
 \medskip
 
\noindent\textbf{\textbf{Qualitative results.}} In figure \ref{fig:4}, we show different kinds of spatial image conditions. Firstly, we visualize the results generated using different types of  conditioned images and text prompts and compare them with mainstream methods. We find that the generated images are able to align closely with the spatial image conditions in different conditional modalities, which implies that our method extracts structure features that are not restricted by modality. Our feature injection is effective for structure guidance and is expected to generalize to other types of control images. In terms of appearance, our method generates visually better and more realistic images. Other methods suffer from distortions when some details are processed (e.g., the number of hands and feet is incorrect or shadows are given colors). This proves that appearance guidance plays an active role in our approach. We also compare it with other layout-based generation methods. In figure \ref{fig:5}, we show the comparison results with GLIGEN\cite{Alpher05}. We can find that GLIGEN is prone to problems of collapsing spatial structure and incorrect attributes (e.g., color), while our images are generally of higher quality. However, we are also prone to face distortion when dealing with sharp object shapes (e.g., seabird beaks). This may be caused by the fact that polygons do not easily fit sharp objects.

\subsection{Ablation Study}
\begin{table}
  \centering
  \begin{tabular}{@{}lcc@{}}
    \toprule
    Methods & $FID \downarrow$ & $CLIP \downarrow$ \\
    \midrule
    ours & 8.44 & 0.3222 \\
    w/o mask & 8.46 & 0.2911 \\
    w/o concatenation & 8.56 & 0.3011 \\ 
    \bottomrule
  \end{tabular}
  \caption{Ablation studies on each component of our method.}
  \label{tab:3}
\end{table}
\textbf{\textbf{Masks on visual features.}} When we remove the mask on visual features, each visual primitive, consisting of an object's layout and appearance description, will not only focus on the corresponding local position, but on the global. We find that the CLIP score decreases more after removing the mask, while the FID remains almost unchanged, which implies that the design of the mask facilitates the enhancement of the layout and structural control over the image with less impact on the visual quality.
\medskip

\noindent\textbf{\textbf{Connection of text embedding to location embedding}}. Instead of connecting the text embedding describing the appearance of the object to the location embedding, we pass the text embedding directly into our masked cross-attention. We found that both FID score and CLIP score decreased, which means that this connection helps to improve the controllability and visual quality to some extent.
\medskip

\noindent\textbf{\textbf{Diffusion feature extraction.}} We injected the self-attention weights of \({I_S}\) directly into the diffusion process of \(I\) without feature extraction with DDIM Inversion. We find that this leads to severe appearance leakage. This means that 
 the extraction of diffusion features is necessary.
\section{Conclusions}

In this work, we have utilized MLLM as a global controller to arrange spatial layouts, enhance semantic descriptions, and bind object properties. Specifically, we first proposed a new visual representation that generates polygonal bounding boxes  bounded with textual outer appearance descriptions for each object in space based on path clipping. By masking cross-attention, we make each visual primitive focus on local locations. We use spatial image conditioning in a feature-injected manner, which enables complementary control of text and image. Our approach achieves competitive results in terms of consistency between image and text, layout alignment, and visual quality.
\medskip

\noindent\textbf{\textbf{Limitations.}} Firstly, the number of vertices is an important metric when fitting polygons to different objects. We currently lack sufficient research to find the optimal design method for the number of vertices. In future work, we will explore the effect of vertex number on model training, inference and generation to set a better vertex number. In addition, for the feature injection session, we need to perform DDIM Inversion as well as feature extraction. This definitely complicates the inference process and increases the computational cost. In the future, we will try more efficient feature extraction and injection methods.
{
    \small
    \bibliographystyle{ieeenat_fullname}
    \bibliography{main}
}


\end{document}